\documentclass[11pt,a4paper]{article}
\usepackage[hyperref]{emnlp2018}
\usepackage{times}
\usepackage{latexsym}
\usepackage[justification=centering]{caption}
\usepackage{graphicx}
\usepackage{tabularx}
\usepackage{soul}
\usepackage{algorithm,algorithmic}
\usepackage{textgreek}
\usepackage{multirow}
\usepackage{enumitem}
\usepackage{tabu}
\usepackage{array}
\usepackage{booktabs}
\usepackage{mathtools}
\usepackage{tikz}
\usepackage[normalem]{ulem}
\usepackage{gensymb}
\usepackage{amssymb}
\usepackage{pifont}
\usepackage{tikz}
\usepackage[utf8x]{inputenc}
\usepackage[export]{adjustbox}
\setlist{noitemsep}
\usepackage{url}

\aclfinalcopy 

\title{Language Identification in Code-Mixed Data using Multichannel Neural Networks and Context Capture}

\author{Soumil Mandal \\
  Department of CSE \\
  SRM University, Chennai, India \\
  {\tt soumil.mandal@gmail.com} \\\And
  Anil Kumar Singh \\
  Department of CSE \\
  IIT (BHU), Varanasi, India \\
  {\tt aksingh.cse@iitbhu.ac.in} \\}

\date{}

\begin{document}
\maketitle

\begin{abstract}
An accurate language identification tool is an absolute necessity for building complex NLP systems to be used on code-mixed data. Lot of work has been recently done on the same, but there's still room for improvement. Inspired from the recent advancements in neural network architectures for computer vision tasks, we have implemented multichannel neural networks combining CNN and LSTM for word level language identification of code-mixed data. Combining this with a Bi-LSTM-CRF context capture module, accuracies of 93.28\% and 93.32\% is achieved on our two testing sets.
\end{abstract}

\section{Introduction}
With the rise of social media, the amount of mineable data is rising rapidly. Countries where bilingualism is popular, we see users often switch back and forth between two languages while typing, a phenomenon known as code-mixing or code-switching. For analyzing such data, language tagging acts as a preliminary step and its accuracy and performance can impact the system results to a great extent. Though a lot of work has been done recently targeting this task, the problem of language tagging in code-mixed scenario is still far from being solved. Code-mixing scenarios where one of the languages have been typed in its transliterated from possesses even more challenges, especially due to inconsistent phonetic typing. On such type of data, context capture is extremely hard as well. Proper context capture can help in solving problems like ambiguity, that is word forms which are common to both the languages, but for which, the correct tag can be easily understood by knowing the context. An additional issue is a lack of available code-mixed data. Since most of the tasks require supervised models, the bottleneck of data crisis affects the performance quite a lot, mostly due to the problem of over-fitting.

\noindent In this article, we present a novel architecture, which captures information at both word level and context level to output the final tag. For word level, we have used a multichannel neural network (MNN) inspired from the recent works of computer vision. Such networks have also shown promising results in NLP tasks like sentence classification \cite{kim2014convolutional}. For context capture, we used Bi-LSTM-CRF. The context module was tested more rigorously as in quite a few of the previous work, this information has been sidelined or ignored. We have experimented on Bengali-English (Bn-En) and Hindi-English (Hi-En) code-mixed data. Hindi and Bengali are the two most popular languages in India. Since none of them have Roman as their native script, both are written in their phonetically transliterated from when code-mixed with English.

\section{Related Work}
In the recent past, a lot of work has been done in the field of code-mixing data, especially language tagging. \citet{king:2013labeling} used weakly semi-supervised methods for building a world level language identifier.  Linear chain CRFs with context information limited to bigrams was employed by \citet{nguyen2013word}. Logistic regression along with a module which gives code-switching probability was used by \citet{vyas2014pos}. Multiple features like word context, dictionary, n-gram, edit distance were used by \citet{das2014identifying}. \citet{jhamtani2014word} combined two classifiers into an ensemble model for Hindi-English code-mixed LID. The first classifier used modified edit distance, word frequency and character n-grams as features. The second classifier used the output of the first classifier for the current word, along with language tag and POS tag of neighboring to give the final tag. \citet{piergallini2016word} made a word level model taking char n-grams and capitalization as feature. \citet{rijhwani2017estimating} presented a generalized language tagger for arbitrary large set of languages which is fully unsupervised. \citet{choudhury-EtAl:2017:W17-75} used a model which concatenated word embeddings and character embeddings to predict the target language tag. \citet{mandal:2018language} used character embeddings along with phonetic embeddings to build an ensemble model for language tagging.

\section{Data Sets}
We wanted to test our approach on two different language pairs, which were Bengali-English (Bn-En) and Hindi-English (Hi-En). For Bn-En, we used the data prepared in \citet{mandal:2018preparing} and for Hi-En, we used the data prepared in \citet{patra:2018sentiment}. The number of instances of each type we selected for our experiments was 6000. The data distribution for each type is shown in Table~\ref{table1}.

\begin{table}[h]
\centering
\begin{tabular}{|l|c|c|c|}
\hline
\multicolumn{1}{|c|}{\textbf{}} & \textbf{Train} & \textbf{Dev} & \textbf{Test} \\ \hline
\textbf{Bn} & \begin{tabular}[c]{@{}c@{}}3000\\ 27245/6189\\ 22.4\end{tabular} & \begin{tabular}[c]{@{}c@{}}1000\\ 9144/2836\\ 21.4\end{tabular} & \begin{tabular}[c]{@{}c@{}}2000\\ 17967/4624\\ 22.5\end{tabular} \\ \hline
\textbf{Hi} & \begin{tabular}[c]{@{}c@{}}3000\\ 26384/5630\\ 18.8\end{tabular} & \begin{tabular}[c]{@{}c@{}}1000\\ 8675/2485 \\ 18.7\end{tabular} & \begin{tabular}[c]{@{}c@{}}2000\\ 16114/4286\\ 18.2\end{tabular} \\ \hline
\end{tabular}
\caption{Data distribution.}
\label{table1}
\end{table}

\noindent Here, the first value represents the number of instances taken, the second line represents the total number of indic tokens / unique number of indic tokens, and the third line represents the mean code-mixing index \cite{das2014identifying}.

\section{Architecture Overview}
Our system is comprised of two supervised modules. The first one is a multichannel neural network trained at word level, while the second one is a simple bidirectional LSTM-CRF trained at instance level. The second module takes the input from the first module along with some other features to output the final tag. Individual modules are described in detail below.
\begin{figure}[h]
\centering
\includegraphics[scale=0.32]{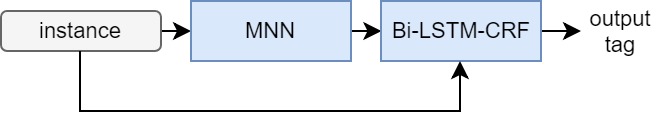}
\caption{Architecture overview.}
\label{fig1}
\end{figure}

\section{Word - Multichannel Neural Network}
\begin{figure}[h]
\centering
\includegraphics[scale=0.38]{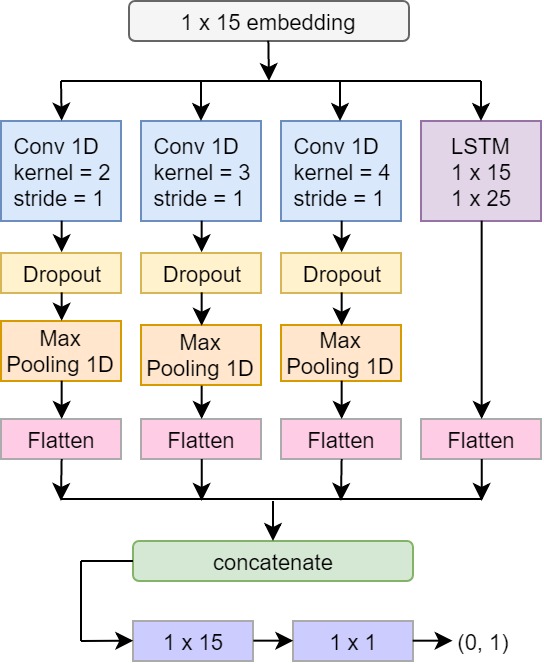}
\caption{Multichannel architecture for word level tagging.}
\label{fig2}
\end{figure}

Inspired from the recent deep neural architectures developed for image classification tasks, especially the Inception architecture \cite{szegedy:2015going}, we decided to use a very similar concept for learning language at word level. This is because the architecture allows the network to capture representations of different types, which can be really helpful for NLP tasks like these as well. The network we developed has 4 channels, the first three enters into a Convolution 1D (Conv1D) network \cite{lecun1999object}, while the fourth one enters into a Long Short Term Memory (LSTM) network \cite{hochreiter1997long}. The complete architecture is showed in Fig ~\ref{fig2}.

\noindent Character embeddings of length 15 is fed into all the 4 channels. The first 3 Conv 1D cells are used to capture n-gram representations. All the three Conv 1D cells are followed by Dropout (rate 0.2) and Max Pooling cells. Adding these cells help in controlling overfitting and learning invariances, as well as reduce computation cost. Activation function for all the three Conv 1D nets was relu. The fourth channel goes to an LSTM stack with two computational layers of sizes 15, and 25 orderly. For all the four channels, the final outputs are flattened and concatenated. This concatenated vector is then passed on to two dense layers of sizes 15 (activation relu) and 1 (activation sigmoid). For the two models created, Bn-En and Hi-En, target labels were 0 for the Bn/Hi and 1 for En. For implementing the multichannel neural network for word level classification, we used the Keras API \cite{chollet:2015keras}.   

\subsection{Training}
Word distribution for training is described in Table~\ref{table1}. All indic tagged tokens were used instead of just unique ones of respective languages. The whole model was compiled using loss as binary cross-entropy, optimization function used was adam \cite{kingma2014adam} and metric for training was accuracy. The batch size was set to 64, and number of epochs was set to 30. Other parameters were kept at default. The training accuracy and loss graphs for both Bn and Hi are shown below. As the MNN model produces a sigmoid output, to convert the model into a classifier, we decided to use a threshold based technique identical to the one used in \citet{mandal:2018language} for tuning character and phonetic models. For this the development data was used, threshold for Bn was calculated to be $\theta$ $\leq$ 0.95, while threshold for Hi was calculated to be $\theta$ $\leq$ 0.89. Using these, the accuracies on the development data was 93.6\% and 92.87\% for Bn and Hi respectively.

\section{Context - Bi-LSTM-CRF}
The purpose of this module is to learn representation at instance level, i.e. capture context. For this, we decided to use bidirectional LSTM network with CRF layer (Bi-LSTM-CRF) \cite{huang2015bidirectional} as it has given state of the art results for sequence tagging in the recent past. For converting the instances into embeddings, two features were used namely, sigmoid output from MNN (\textit{fe1}), character embedding (\textit{fe2}) of size 30. The final feature vector is created by concatenating these two, \textit{fe} = (\textit{fe1}, \textit{fe2}). The model essentially learns code-switching chances or probability taking into consideration character embeddings and sigmoid scores of language tag. We used the open sourced neural sequence labeling toolkit, NCRF++ \cite{yang2018ncrf} for building the model. 

\subsection{Training}
Instance distribution for training is described in Table~\ref{table1}. The targets here were the actual language labels (0 for the Bn/Hi and 1 for En). The hyper-parameters which we set mostly follow \citet{yang2018design} and \citet{ma2016end}. Both the models (Bn-En \& Hi-En) had identical parameters. \textit{L}\textsubscript{2} regularization $\lambda$ was set at 1e-8. Learning rate $\eta$ was set to 0.015. Batch size was kept at 16 and number of epochs was set to 210. Mini-batch stochastic gradient descent (SGD) with decaying learning rate (0.05) was used to update the parameters. All the other parameters were kept at default values. This setting was finalized post multiple experiments on the development data. Final accuracy scores on the development data was 93.91\% and 93.11\% for Bn and Hi respectively.  

\section{Evaluation}
For comparison purposes, we decided to use the character encoding architecture described in \citet{mandal:2018language} (stacked LSTMs of sizes 15, 35, 25, 1) with identical hyper-parameters for both Bn and Hi. Training data distribution while creating the baseline models were in accordance with Table~\ref{table1}. The thresholds for the baseline models calculated on the development data was found to be $\theta$ $\leq$ 0.91 and $\theta$ $\leq$ 0.90 for Bn and Hi respectively. The results (in \%) for each of the language pairs are shown below. 

\begin{table}[H]
\centering
\begin{tabular}{|l|c|c|c|c|}
\hline
\multicolumn{1}{|c|}{\textbf{}} & \textbf{Acc} & \textbf{Prec} & \textbf{Rec} & \textbf{F1} \\ \hline
\textit{baseline} &88.32 &89.64  &87.72  &88.67  \\ \hline
\textit{word model} &92.87 &94.33  &91.84  &93.06  \\ \hline
\textit{context model} & \textbf{93.28} &94.33  &92.68  &93.49  \\ \hline
\end{tabular}
\caption{Evaluation on Bn.}
\label{table2}
\end{table}

\noindent From Table~\ref{table2} we can see that the jump in accuracy from baseline to the word model is quite significant (4.55\%). From word to context model, though not much, but still an improvement is seen (0.41\%).
\begin{table}[H]
\centering
\begin{tabular}{|l|c|c|c|c|}
\hline
\multicolumn{1}{|c|}{\textbf{}} & \textbf{Acc} & \textbf{Prec} & \textbf{Rec} & \textbf{F1} \\ \hline
\textit{baseline} &88.28  &88.57  &88.01  &88.28  \\ \hline
\textit{word model} &92.65  &93.54  &91.77  &92.64  \\ \hline
\textit{context model} & \textbf{93.32} &93.62  &93.03  &93.32  \\ \hline
\end{tabular}
\caption{Evaluation on Hi.}
\label{table3}
\end{table}
\noindent In Table~\ref{table3}, again a similar pattern can be seen, i.e. a significant improvement (4.37\%) from baseline to word model. Using the context model, accuracy increases by 0.67\%. In both the Tables, we see that precision has been much higher than recall. 

\section{Analysis \& Discussion}
The confusion matrices of the language tagging models are shown in Table~\ref{table4} and Table~\ref{table5} for Bn and Hi respectively. Predicted class is denoted by Italics, while Roman shows the True classes.

\newcolumntype{?}{!{\vrule width 1.3pt}}
\begin{table}[H]
\centering
\begin{tabular}{|c|c|c?c|c|c|}
\hline
\multicolumn{6}{|c|}{\textbf{Confusion Matrices}} \\ \hline
\multicolumn{1}{|c|}{1} & Bn & En & 2 & Bn & En \\ \hline
\textit{Bn} & 16502 & 1465 & \textit{Bn} & 16652 & 1315 \\ \hline
\textit{En} & 991 & 15521 & \textit{En} & 1000 & 15512 \\ \hline
\end{tabular}
\caption{Confusion matrices for Bn.}
\label{table4}
\end{table}

\noindent From Table~\ref{table4} (1 - word model, 2 - context model), we can see that the correctly predicted En tokens has not varied much, but in case of Bn, the change is quite substantial, and the accuracy improvement from word to context model is contributed by this. Upon analyzing the tokens which were correctly classified by context model, but misclassified by word model, we see that most of them are rarely used Bn words, e.g. \textit{shaaotali} (tribal), \textit{lutpat} (looted), \textit{golap} (rose), etc, or words with close phonetic similarity with an En word(s) or with long substrings which belong to the En vocabulary, e.g. \textit{chata} (umbrella), \textit{botol} (bottle), \textit{gramin} (rural), etc. For some instances, we do see that ambiguous words have been correctly tagged by the context model unlike the word model, where the same language tag is given.

\begin{flushleft}
\textit{E.g 1.} Amar\textbackslash bn shob\textbackslash bn rokom\textbackslash bn er\textbackslash bn e\textbackslash bn fruit\textbackslash en like\textbackslash en aam\textbackslash bn, jam\textbackslash bn, kathal\textbackslash bn bhalo\textbackslash bn lage\textbackslash bn. (Trans. I like all kinds of fruits like aam, jam, kathal.)
\end{flushleft}

\begin{flushleft}
\textit{E.g 2.} Sath\textbackslash bn shokale\textbackslash bn eto\textbackslash bn jam\textbackslash en eriye\textbackslash bn office\textbackslash en jawa\textbackslash bn is\textbackslash en a\textbackslash en big\textbackslash en headache\textbackslash en amar\textbackslash bn boyeshe\textbackslash bn. (Trans. Early morning commuting through traffic for office is a big headache at my age.)
\end{flushleft}

\noindent In the first example, the word "jam" is a Bengali word meaning rose apple (a type of tropical fruit), while in the second example, the word "jam" is an English word referring to traffic jam i.e. traffic congestion. Thus, we can see that despite having same spellings, the word has been classified to different languages, and that too correctly. This case was observed in 47 instances, while for 1 instance, it tagged the ambiguous word incorrectly. Thus we see that when carefully trained on standard well annotated data, the positive impact is much higher than negative. 

\noindent In Table~\ref{table5} (3 - word model, 4 - context model) we can see substantial improvement in prediction of En tokens as well by the context model, though primary reason for accuracy improvement is due to better prediction of Hi tokens.
\vspace*{0.18cm}
\begin{table}[H]
\centering
\begin{tabular}{|c|c|c?c|c|c|}
\hline
\multicolumn{6}{|c|}{\textbf{Confusion Matrices}} \\ \hline
\multicolumn{1}{|c|}3 & Hi & En & 4
& Hi & En \\ \hline
\textit{Hi} & 14788 & 1326 & \textit{Hi} & 14992 & 1122 \\ \hline
\textit{En} & 1034 & 14968 & \textit{En} & 1021 & 14981 \\ \hline
\end{tabular}
\caption{Confusion matrices for Hi.}
\label{table5}
\end{table}

\noindent Here again, on analyzing the correct predictions by the context model but misclassified by the word model, we see a similar pattern of rarely used Hi words, e.g. \textit{pasina} (sweat), \textit{gubare} (balloon), \textit{} or Hi words which have phonetic similarities with En words, e.g. \textit{tabla} (a musical instrument), \textit{jangal} (jungle), \textit{pajama} (pyjama), etc. In the last two cases, we can see that the words are actually borrowed words. Some ambiguous words were correctly tagged here as well.

\begin{flushleft}
\textit{E.g 3.} First\textbackslash en let\textbackslash en me\textbackslash en check\textbackslash en fir\textbackslash hi age\textbackslash hi tu\textbackslash hi deklena\textbackslash hi. (Trans. First let me check then later you takeover.)
\end{flushleft}

\begin{flushleft}
\textit{E.g 4.} Anjan\textbackslash hi woman\textbackslash en se\textbackslash hi age\textbackslash en puchna\textbackslash hi is\textbackslash en wrong\textbackslash en. (Trans. Asking age from an unknown woman is wrong.)
\end{flushleft}

\noindent In the first example, "age" is a Hindi word meaning ahead, while in the next instance, "age" is an English word meaning time that a person has lived. Here, correct prediction for ambiguous words was seen in 39 instances while there was no wrong predictions.

\section{Conclusion \& Future Work}
In this article, we have presented a novel architecture for language tagging of code-mixed data which captures context information. Our system achieved an accuracy of 93.28\% on Bn data and 93.32\% on Hi data. The multichannel neural network alone got quite impressive scores of 92.87\% and 92.65\% on Bn and Hi data respectively. In future, we would like to incorporate borrowed (\citet{hoffer2002language}, \citet{haspelmath2009lexical}) tag and collect even more code-mixed data for building better models. We would also like to experiment with variants of the architecture shown in Fig~\ref{fig1} on other NLP tasks like text classification, named entity recognition, etc.    

\bibliography{emnlp2018}

\begin{thebibliography}{23}
\expandafter\ifx\csname natexlab\endcsname\relax\def\natexlab#1{#1}\fi

\bibitem[{Chollet et~al.(2015)}]{chollet:2015keras}
Fran\c{c}ois Chollet et~al. 2015.
\newblock Keras.
\newblock \url{https://keras.io}.

\bibitem[{Choudhury et~al.(2017)Choudhury, Bali, Sitaram, and
  Baheti}]{choudhury-EtAl:2017:W17-75}
Monojit Choudhury, Kalika Bali, Sunayana Sitaram, and Ashutosh Baheti. 2017.
\newblock Curriculum design for code-switching: Experiments with language
  identification and language modeling with deep neural networks.
\newblock In \emph{Proceedings of the 14th International Conference on Natural
  Language Processing (ICON-2017)}, pages 65--74, Kolkata, India. NLP
  Association of India.

\bibitem[{Das and Gamb{\"a}ck(2014)}]{das2014identifying}
Amitava Das and Bj{\"o}rn Gamb{\"a}ck. 2014.
\newblock Identifying languages at the word level in code-mixed indian social
  media text.

\bibitem[{Haspelmath(2009)}]{haspelmath2009lexical}
Martin Haspelmath. 2009.
\newblock Lexical borrowing: Concepts and issues.
\newblock \emph{Loanwords in the world’s languages: A comparative handbook},
  pages 35--54.

\bibitem[{Hochreiter and Schmidhuber(1997)}]{hochreiter1997long}
Sepp Hochreiter and J{\"u}rgen Schmidhuber. 1997.
\newblock Long short-term memory.
\newblock \emph{Neural computation}, 9(8):1735--1780.

\bibitem[{Hoffer(2002)}]{hoffer2002language}
Bates~L Hoffer. 2002.
\newblock Language borrowing and language diffusion: An overview.
\newblock \emph{Intercultural communication studies}, 11(4):1--37.

\bibitem[{Huang et~al.(2015)Huang, Xu, and Yu}]{huang2015bidirectional}
Zhiheng Huang, Wei Xu, and Kai Yu. 2015.
\newblock Bidirectional lstm-crf models for sequence tagging.
\newblock \emph{arXiv preprint arXiv:1508.01991}.

\bibitem[{Jhamtani et~al.(2014)Jhamtani, Bhogi, and
  Raychoudhury}]{jhamtani2014word}
Harsh Jhamtani, Suleep~Kumar Bhogi, and Vaskar Raychoudhury. 2014.
\newblock Word-level language identification in bi-lingual code-switched texts.
\newblock In \emph{Proceedings of the 28th Pacific Asia Conference on Language,
  Information and Computing}.

\bibitem[{Kim(2014)}]{kim2014convolutional}
Yoon Kim. 2014.
\newblock Convolutional neural networks for sentence classification.
\newblock \emph{arXiv preprint arXiv:1408.5882}.

\bibitem[{King and Abney(2013)}]{king:2013labeling}
Ben King and Steven Abney. 2013.
\newblock Labeling the languages of words in mixed-language documents using
  weakly supervised methods.
\newblock In \emph{Proceedings of the 2013 Conference of the North American
  Chapter of the Association for Computational Linguistics: Human Language
  Technologies}, pages 1110--1119.

\bibitem[{Kingma and Ba(2014)}]{kingma2014adam}
Diederik~P Kingma and Jimmy Ba. 2014.
\newblock Adam: A method for stochastic optimization.
\newblock \emph{arXiv preprint arXiv:1412.6980}.

\bibitem[{LeCun et~al.(1999)LeCun, Haffner, Bottou, and
  Bengio}]{lecun1999object}
Yann LeCun, Patrick Haffner, L{\'e}on Bottou, and Yoshua Bengio. 1999.
\newblock Object recognition with gradient-based learning.
\newblock In \emph{Shape, contour and grouping in computer vision}, pages
  319--345. Springer.

\bibitem[{Ma and Hovy(2016)}]{ma2016end}
Xuezhe Ma and Eduard Hovy. 2016.
\newblock End-to-end sequence labeling via bi-directional lstm-cnns-crf.
\newblock \emph{arXiv preprint arXiv:1603.01354}.

\bibitem[{Mandal et~al.(2018{\natexlab{a}})Mandal, Das, and
  Das}]{mandal:2018language}
Soumil Mandal, Sourya~Dipta Das, and Dipankar Das. 2018{\natexlab{a}}.
\newblock Language identification of bengali-english code-mixed data using
  character \& phonetic based lstm models.
\newblock \emph{arXiv preprint arXiv:1803.03859}.

\bibitem[{Mandal et~al.(2018{\natexlab{b}})Mandal, Mahata, and
  Das}]{mandal:2018preparing}
Soumil Mandal, Sainik~Kumar Mahata, and Dipankar Das. 2018{\natexlab{b}}.
\newblock Preparing bengali-english code-mixed corpus for sentiment analysis of
  indian languages.
\newblock \emph{arXiv preprint arXiv:1803.04000}.

\bibitem[{Nguyen and Do{\u{g}}ru{\"o}z(2013)}]{nguyen2013word}
Dong Nguyen and A~Seza Do{\u{g}}ru{\"o}z. 2013.
\newblock Word level language identification in online multilingual
  communication.
\newblock In \emph{Proceedings of the 2013 Conference on Empirical Methods in
  Natural Language Processing}, pages 857--862.

\bibitem[{Patra et~al.(2018)Patra, Das, and Das}]{patra:2018sentiment}
Braja~Gopal Patra, Dipankar Das, and Amitava Das. 2018.
\newblock Sentiment analysis of code-mixed indian languages: An overview of
  sail\_code-mixed shared task@ icon-2017.
\newblock \emph{arXiv preprint arXiv:1803.06745}.

\bibitem[{Piergallini et~al.(2016)Piergallini, Shirvani, Gautam, and
  Chouikha}]{piergallini2016word}
Mario Piergallini, Rouzbeh Shirvani, Gauri~S Gautam, and Mohamed Chouikha.
  2016.
\newblock Word-level language identification and predicting codeswitching
  points in swahili-english language data.
\newblock In \emph{Proceedings of the Second Workshop on Computational
  Approaches to Code Switching}, pages 21--29.

\bibitem[{Rijhwani et~al.(2017)Rijhwani, Sequiera, Choudhury, Bali, and
  Maddila}]{rijhwani2017estimating}
Shruti Rijhwani, Royal Sequiera, Monojit Choudhury, Kalika Bali, and
  Chandra~Shekhar Maddila. 2017.
\newblock Estimating code-switching on twitter with a novel generalized
  word-level language detection technique.
\newblock In \emph{Proceedings of the 55th Annual Meeting of the Association
  for Computational Linguistics (Volume 1: Long Papers)}, volume~1, pages
  1971--1982.

\bibitem[{Szegedy et~al.(2015)Szegedy, Liu, Jia, Sermanet, Reed, Anguelov,
  Erhan, Vanhoucke, Rabinovich et~al.}]{szegedy:2015going}
Christian Szegedy, Wei Liu, Yangqing Jia, Pierre Sermanet, Scott Reed, Dragomir
  Anguelov, Dumitru Erhan, Vincent Vanhoucke, Andrew Rabinovich, et~al. 2015.
\newblock Going deeper with convolutions.
\newblock Cvpr.

\bibitem[{Vyas et~al.(2014)Vyas, Gella, Sharma, Bali, and
  Choudhury}]{vyas2014pos}
Yogarshi Vyas, Spandana Gella, Jatin Sharma, Kalika Bali, and Monojit
  Choudhury. 2014.
\newblock Pos tagging of english-hindi code-mixed social media content.
\newblock In \emph{Proceedings of the 2014 Conference on Empirical Methods in
  Natural Language Processing (EMNLP)}, pages 974--979.

\bibitem[{Yang et~al.(2018)Yang, Liang, and Zhang}]{yang2018design}
Jie Yang, Shuailong Liang, and Yue Zhang. 2018.
\newblock Design challenges and misconceptions in neural sequence labeling.
\newblock \emph{arXiv preprint arXiv:1806.04470}.

\bibitem[{Yang and Zhang(2018)}]{yang2018ncrf}
Jie Yang and Yue Zhang. 2018.
\newblock Ncrf++: An open-source neural sequence labeling toolkit.
\newblock In \emph{Proceedings of the 56th Annual Meeting of the Association
  for Computational Linguistics}.

\end{thebibliography}
\bibliographystyle{acl_natbib_nourl}

\end{document}